\documentclass[letterpaper, 10 pt, conference]{ieeeconf}  

\IEEEoverridecommandlockouts                              

\usepackage{algorithm}
\usepackage{algpseudocode}
\usepackage{graphicx}
\usepackage{subcaption}
\usepackage{tcolorbox}
\usepackage{booktabs}
\usepackage{amsmath}
\usepackage{amssymb}
\let\labelindent\relax
\usepackage[inline]{enumitem}
 \usepackage{xspace}
\usepackage{siunitx}
\usepackage{adjustbox}
\usepackage[caption=false,font=footnotesize]{subfig}

\usepackage{makecell}  
\usepackage[table,dvipsnames]{xcolor}
\usepackage[maxbibnames=1,minbibnames=1,backend=biber,sorting=nty]{biblatex}
\newcommand{\modelname}{\textsc{NeuroSymLand}\xspace}

\title{\LARGE \bf
\modelname: Neuro-Symbolic Landing-Site Assessment for Robust and Edge-Deployable UAV Autonomy

}

\author{
Weixian Qian$^{1,*}$,
Tianyi Yang$^{2,*}$,
and Sebastian Schroder$^{1,*}$\\
Yao Deng$^{1,3}$,
Jiaohong Yao$^{1}$,
Xiao Cheng$^{1}$,
Richard Han$^{1}$,
and Xi Zheng$^{1,\dagger}$%
\thanks{$^{*}$Equal contribution.}%
\thanks{$^{\dagger}$Corresponding author.}%
\thanks{$^{1}$School of Computing, Macquarie University, Sydney, NSW, Australia.
{\tt\small \{weixian.qian, sebastian.schroder, jiaohong.yao, xiao.cheng, richard.han, james.zheng\}@mq.edu.au}}%
\thanks{$^{3}$Yao Deng was with Macquarie University, when this work was conducted, and is now with Anitron.
{\tt\small yao.deng@anitron.com}}%
\thanks{$^{2}$Department of Computer Science, University of California, Santa Barbara, CA, USA.
{\tt\small tianyi\_yang@ucsb.edu}}%
}

\begin{document}

\maketitle
\thispagestyle{empty}
\pagestyle{empty}

\begin{abstract}
Safe landing-site assessment in unstructured environments remains a key challenge for autonomous UAV deployment, as vision-only learning approaches often degrade under terrain variability and provide limited transparency in safety decisions. We present \modelname, a neuro-symbolic landing-site assessment system that integrates lightweight perception with explicit safety reasoning. The framework constructs a probabilistic semantic scene graph from onboard visual input and evaluates candidate landing regions using symbolic constraints capturing terrain flatness, obstacle clearance, and spatial consistency, enabling structured reasoning under perceptual uncertainty while maintaining edge-feasible execution. Across 72 simulated landing scenarios spanning diverse terrains, \modelname achieves 61 successful assessments, outperforming four competitive baselines (37-57 successes). To evaluate deployability, we further conduct 100 hardware-in-the-loop trials with randomized initial poses, profiling end-to-end latency, stage-wise execution time, and system-level metrics including CPU/GPU utilization, memory footprint, and power consumption. 
Results demonstrate improved robustness and interpretability with bounded edge-resource usage. 
Profiling shows that symbolic reasoning contributes only a small fraction of end-to-end latency, 
while the main computational cost arises from perception and PSSG construction. 
These results demonstrate the feasibility of deploying the landing-site assessment stack on edge-constrained UAV hardware, and all source code, datasets, prompts, and symbolic rule refinement examples are released in an open-source repository\footnote{https://github.com/Janus117/NeuroSymbolicLand}.


\end{abstract}


\section{INTRODUCTION}

Unmanned aerial vehicles (UAVs) are increasingly deployed near people and infrastructure for delivery, inspection, and emergency response, where safe landing remains a persistent bottleneck for reliable field operation. Marker-free \emph{landing-site assessment} (LSA) must select safe touchdown regions under perceptual uncertainty, partial observability, and mission-dependent constraints, where safety depends not only on geometric cues such as flatness and obstacle clearance but also on contextual factors including nearby people, sensitive facilities, environmental affordances, and task-specific policies. Most existing marker-free LSA pipelines~\cite{hinzmann2018freelsd,schoppmann2021multiresolutionelevationmappingsafe,secchiero2024visualenvironmentassessmentsafe,tan2025vislandingmonocular3dperception} rely on end-to-end learning models that regress landing feasibility directly from geometric and semantic features, tightly coupling perception and decision logic. While effective at identifying obvious hazards, such approaches limit interpretability, systematic debugging, and mission reconfiguration, as updating operational constraints often requires retraining or heuristic redesign. Recent LLM/VLM-based methods~\cite{aghaee2025rb,barbosa2024robust,javaid2024large} move toward higher-level reasoning but typically depend on implicit prompting or black-box inference at runtime, complicating deterministic analysis and certification for edge-deployed robotic systems.

\begin{figure}[t]
    \centering
    \includegraphics[width=1.0\columnwidth]{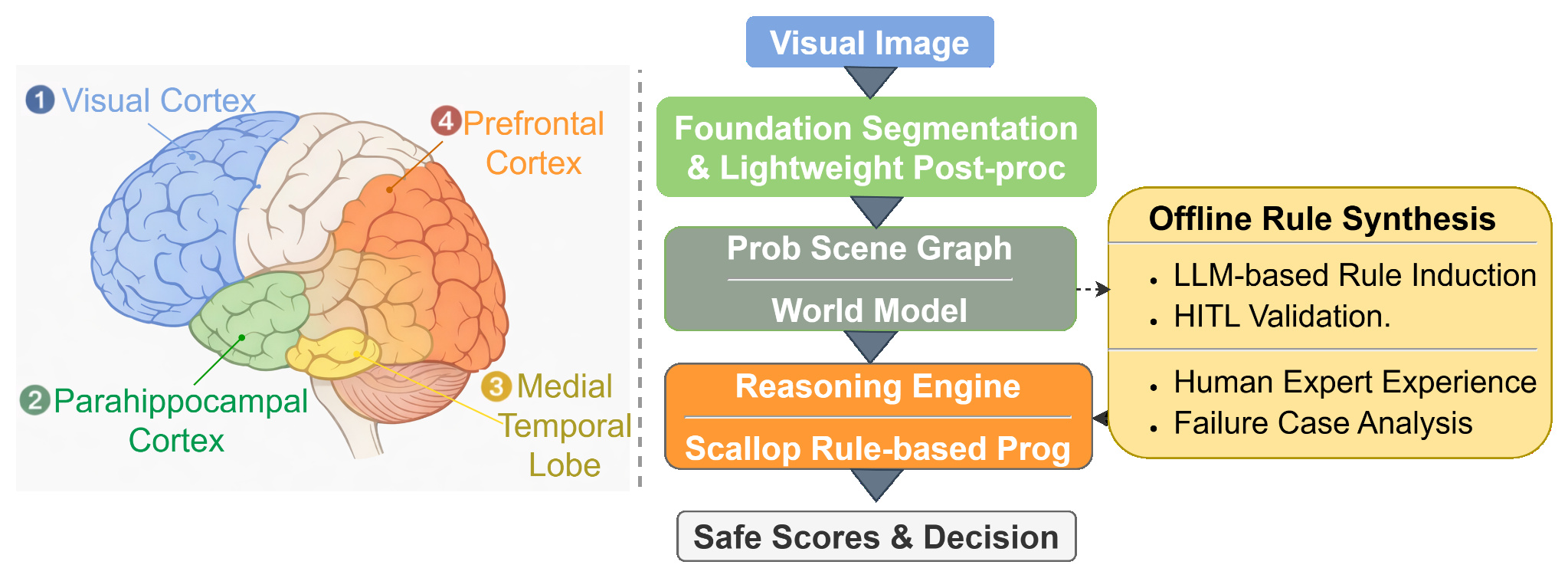}
    \caption{Functional correspondence between brain systems and a perception-world modeling-reasoning architecture with offline cognitive control.}
    \label{fig:brain_vs_model}
\end{figure}

To address these limitations, we propose \modelname{}, a neuro-symbolic LSA framework that explicitly separates perceptual world modeling from mission-level safety reasoning. As illustrated in Figure~\ref{fig:brain_vs_model}, a monocular image is processed by a lightweight segmentation model with geometric post-processing to extract semantic regions and terrain structure, which are assembled into a probabilistic semantic scene graph (\(PSSG\)) serving as an explicit and inspectable world model encoding objects, spatial relations, region attributes, and uncertainty. This structured representation can be directly visualized and audited, enabling systematic debugging of perceptual outputs prior to decision-making. Crucially, the reasoning stage is entirely symbolic: mission constraints are authored offline as explicit logical rules assisted by LLMs during drafting and refined through human-in-the-loop validation and compiled into a deterministic \textsc{Scallop}-based reasoning engine. 
At runtime, no neural inference is involved in landing-region assessment; symbolic rules execute directly over the scene graph to produce traceable safety scores and reasoning paths.
By decoupling perception from fully symbolic decision control, \modelname{} supports structured reasoning under uncertainty while maintaining transparency, debuggability, robustness, and edge deployability, making it well-suited for real-world UAV landing autonomy.
Concretely, this paper makes the following contributions:
\begin{itemize}[leftmargin=17pt,topsep=0pt]
    \item \textbf{Modular neuro-symbolic LSA architecture.} We introduce a landing-site assessment framework that explicitly separates probabilistic world modeling from fully symbolic mission-level reasoning, enabling structured and inspectable decision control under perceptual uncertainty.
    
    \item \textbf{Fully symbolic, deterministic runtime reasoning.} We construct an explicit PSSG from lightweight perception and execute mission constraints as compiled \textsc{Scallop} rules, producing traceable, debuggable, and reconfigurable landing decisions without neural inference in the decision layer.
    
    \item \textbf{LLM-assisted offline rule engineering, no runtime LLM.} We use an LLM only as an offline rule-authoring assistant to translate expert natural-language constraints and failure-case feedback into candidate \textsc{Scallop} rules. The final rule set is inspected, validated, and then executed at runtime without LLM inference.
    
    \item \textbf{System-level evaluation under edge constraints.} Across 72 simulated landing scenarios and 100 hardware-in-the-loop profiling trials, we demonstrate improved landing assessment robustness over competitive baselines while maintaining low latency, bounded memory footprint, and stable embedded execution.
\end{itemize}

\section{Related Work}
\label{sec:related_work}

\subsection{Learning-based landing perception and assessment}
Early landing systems relied on artificial fiducials such as AprilTag~\cite{Olson2011AprilTag,wang2016apriltag2} and ArUco~\cite{GarridoJurado2014ArUco}, but these approaches are restrictive outside controlled environments due to deployment overhead and sensitivity to occlusion and illumination. Consequently, markerless LSA, which infers landing feasibility directly from onboard sensing, has become the dominant paradigm in field robotics. Recent markerless pipelines are largely learning-driven, estimating suitability from terrain geometry (e.g., flatness, slope, roughness, clearance) combined with semantic vetoes (e.g., people, vehicles, water, vegetation)~\cite{hinzmann2018freelsd,schoppmann2021multiresolutionelevationmappingsafe,secchiero2024visualenvironmentassessmentsafe,tan2025vislandingmonocular3dperception}. Other systems integrate elevation mapping, semantic segmentation, stereo, or depth sensing to recover terrain structure and reject unsafe regions~\cite{Chatzikalymnios2022LandingSiteDetection}. While these approaches improve robustness to appearance variation, they typically output a landing score or mask without constructing an explicit representation of scene context beyond the model prediction.


\subsection{Risk-Averse Planning and Landing}
A distinct way of solving this problem is to formulate it as an optimisation problem that balances terrain risk, travel cost, and operational constraints.~\cite{yang2022optimizationlanding} For emergency or time-critical scenarios, systems often integrate mapping with online replanning.~\cite{Zheng2021SafeLandingSiteCONFCDS,Magrisso2021LocalTrajectoryLanding} These approaches, however, typically rely on handcrafted cost maps or task-specific risk models and lack a unified mechanism for expressing higher-level contextual constraints grounded in scene semantics and spatial relations.

\subsection{Hybrid modular approaches: semantic mapping plus explicit decision logic}
Hybrid pipelines maintain structured environmental representations and apply explicit decision logic on top. Semantic SLAM and mapping-based landing systems construct 3D maps augmented with semantic layers (e.g., occupancy, grid, topology) to support landing decisions in unknown environments~\cite{yang2024semanticslamlanding}, while point cloud-based approaches adopt coarse-to-fine strategies that combine geometric assessment with semantic understanding to identify candidate regions~\cite{yang2022pointcloudlanding}. Although these modular designs improve system-level interpretability, their decision modules are often implemented as thresholds, heuristic rankings, or tightly coupled learned scoring. Our approach follows this hybrid paradigm but makes two explicit commitments: we represent the environment as a PSSG rather than only cost maps or masks, and we encode mission logic as explicit symbolic programs executed over the world model. This design aligns with recent neuro-symbolic frameworks such as \textsc{NeuroStrata} and \textsc{Scallop}, which tightly integrate neural perception with symbolic reasoning and support probabilistic logic with inspectable execution traces and efficient inference over structured facts~\cite{zheng2025neurostrata,huang2021scallop,Li2023ScallopPLDI}.

\section{Methodology}
\label{sec:method}



\begin{figure*}[t]
    \centering
    
    \begin{minipage}{0.49\textwidth}
        \centering
        \includegraphics[width=\textwidth]{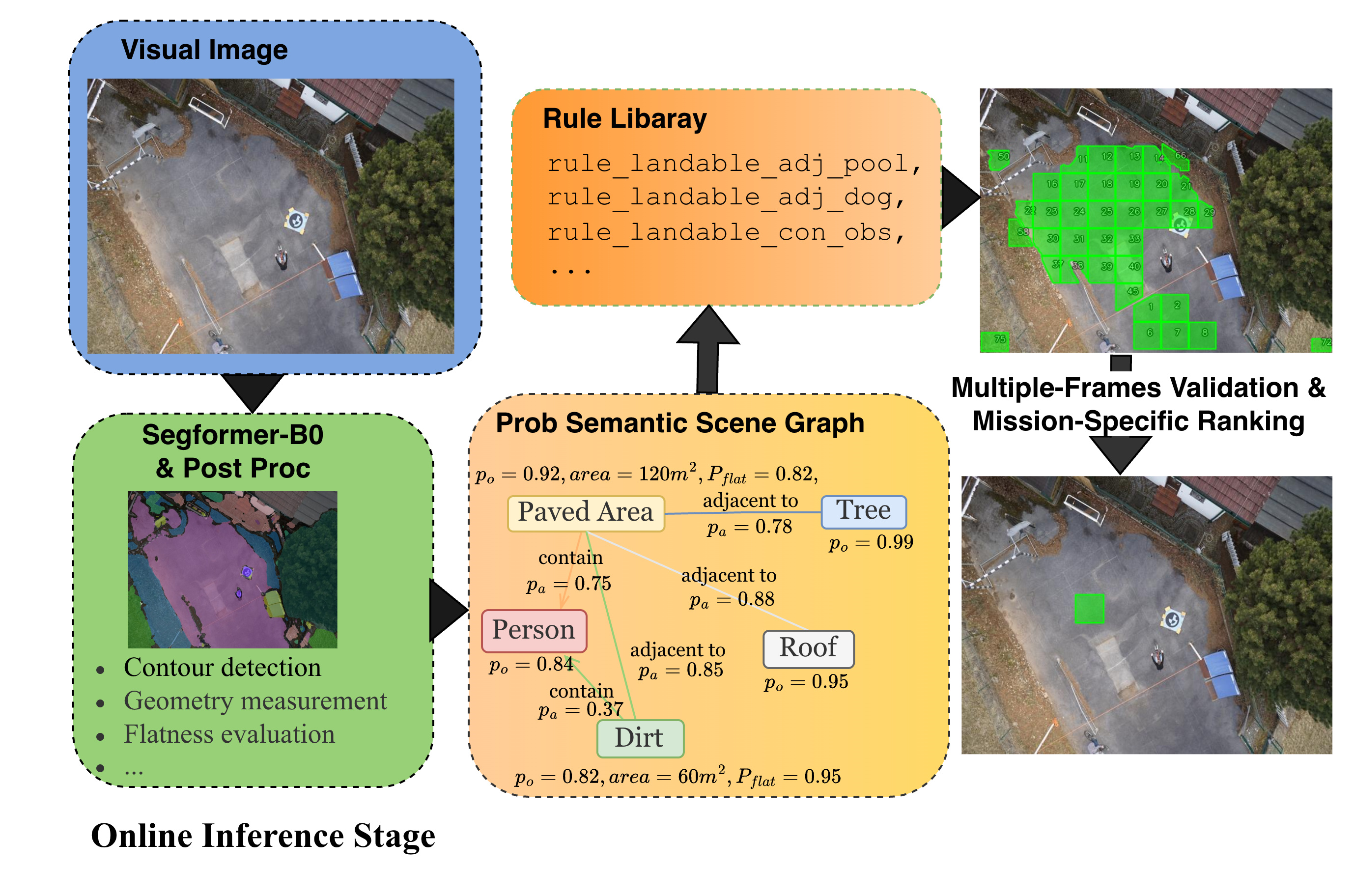}
        \caption{Online inference in \modelname{}. The system segments visual inputs and post-processes them to build a PSSG
 evaluates symbolic safety rules to filter unsafe regions, then applies multi-frame validation and mission-specific ranking to produce interpretable landing decisions.}
        \label{fig:online_inference_stage}
    \end{minipage}
    \hfill
    \begin{minipage}{0.5\textwidth}
        \centering
 \includegraphics[width=\linewidth]{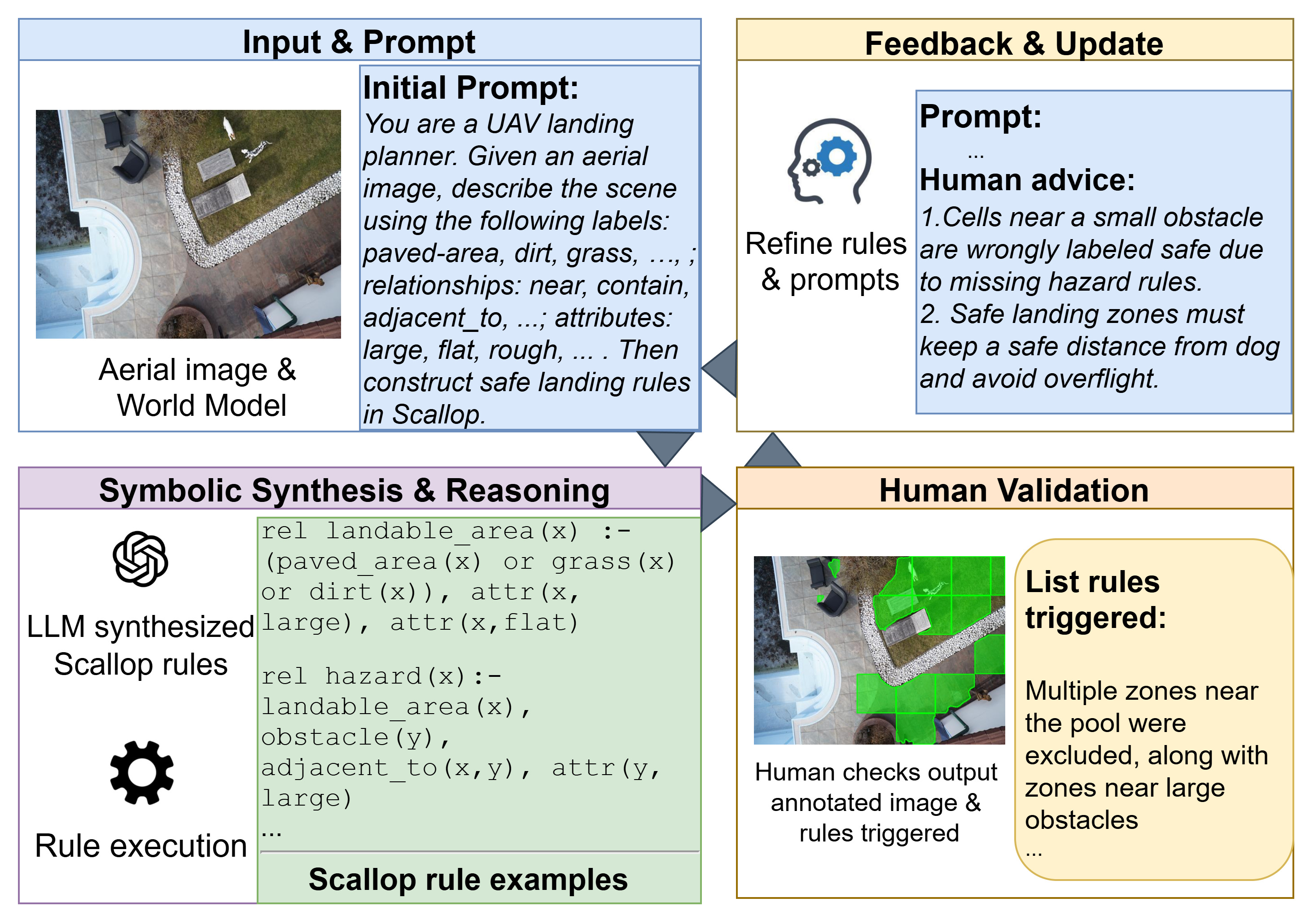}
        \caption{Flow chart for iterative offline rule engine construction with human-in-the-loop refinement.
        Each round updates the symbolic rule library based on accumulated world model evidence and targeted failure case analysis.}
        \label{fig:offline_training}
    \end{minipage}
\end{figure*}
%


\textbf{Overview:}
\modelname{} is a neuro-symbolic LSA framework that separates learned perception, explicit world modeling, and symbolic mission logic. As shown in Fig.~\ref{fig:online_inference_stage}, the online pipeline uses an INT8-quantized SegFormer-B0 backbone with lightweight geometric processing to extract region candidates and attributes. These outputs are lifted into a PSSG serving as the explicit world model for downstream reasoning. A validated \textsc{Scallop} rule set is then executed over the PSSG to infer hazard and safety scores with provenance traces. Candidate regions that pass symbolic filtering are further checked through multi-frame validation and finally ranked using mission-conditioned utility weights to select the landing proposal.

The symbolic rule set is constructed offline, as illustrated in Fig.~\ref{fig:offline_training}. An LLM serves only as a rule-authoring assistant, proposing draft \textsc{Scallop} rules from the fixed PSSG schema, rule libraries, and labeled validation cases. These rules are evaluated in simulation, and experts inspect misclassifications and rule traces to iteratively refine the symbolic rules through prompt updates. This process modifies only the symbolic reasoning layer, while the perception backbone, PSSG schema, and world-model construction remain unchanged. Figure~\ref{fig:offline_training} summarizes the overall workflow; the accompanying online repository\footnote{\url{https://github.com/Janus117/NeuroSymbolicLand}} provides the complete source code, experimental data, prompts, and a step-by-step example of the human-in-the-loop rule refinement process.


We use human-in-the-loop (HITL) to refer to this offline expert review and rule-refinement process. At runtime, no LLM is queried: the system executes a fixed, expert-validated rule set over the current PSSG. The mission score introduced in Sec.~III-C is likewise used only as an inference-time ranking utility after symbolic safety filtering, rather than as a training loss or reward signal. This work focuses on landing-site assessment and edge deployment feasibility, which are necessary prerequisites for full onboard robotic autonomy.

\begin{table}[t]
\vspace{12mm}
\caption{PSSG predicate vocabulary for landing-site assessment. Object predicates are adapted from the Semantic Drone Dataset~\cite{semanticDroneData}; attribute and relation predicates are task-specific.}
\label{tab:scene_graph}
\centering
\begin{tabular}{ll}
\toprule
\textbf{\begin{tabular}[c]{@{}l@{}}Semantic \\ Objects\end{tabular}} &
\begin{tabular}[c]{@{}l@{}}unlabeled, paved area, dirt, grass, gravel,\\ water, rocks, pool, roof, wall, window, door,\\ fence, person, dog, car, bicycle, tree, obstacle\end{tabular} \\
\midrule
\textbf{Attributes} &
\begin{tabular}[c]{@{}l@{}}is large area, is regular shape, is flat surface,\\ is stable surface, is moving, is smooth surface,\\ is accessible\end{tabular} \\
\midrule
\textbf{Relations} &
\begin{tabular}[c]{@{}l@{}}above, below, left, right, adjacent\_to, contain, on,\\ near\_to, far\_from, surrounded\_by\end{tabular} \\
\bottomrule
\end{tabular}
\end{table}

\subsection{Online Generation of Explicit World Models}
\label{subsec:world_model}

The online pipeline transforms raw visual input into an explicit, uncertainty-aware world model that serves as the sole interface to the symbolic reasoning engine. Importantly, safety is \emph{not} inferred by the neural network; instead, perception evidence is deterministically lifted into structured symbolic facts.

\paragraph{Fixed world-model taxonomy}
We define a fixed world-model schema of object classes, attributes, and spatial relations derived from the Semantic Drone Dataset~\cite{semanticDroneData} and landing-specific scene elements (Table~\ref{tab:scene_graph}). The schema is shared by online inference and offline rule construction, mapping all perceptual outputs to auditable symbolic predicates.


\paragraph{Deterministic region and relation construction}
Segmentation posteriors are converted into region instances using connected-component and contour extraction routines~\cite{opencv_library}, with small-region suppression and optional splitting of oversized components. Each region forms a node in the world model and stores semantic confidence together with compact geometric attributes (area, centroid, shape descriptors, and lightweight flatness/slope proxies), optionally temporally smoothed over a short window. Pairwise spatial relations (e.g., \texttt{contain}, \texttt{adjacent\_to}, \texttt{near\_to}) are computed deterministically from region geometry and encoded as probabilistic edges.

\paragraph{Probabilistic Semantic Scene Graph}
The environment is represented as a PSSG $G=(V,E)$, where nodes correspond to segmented regions and edges represent spatial relations. Node attributes store semantic posteriors and geometric features from the fixed taxonomy, while edges encode relation confidences derived from deterministic geometric rules. These values are grounded into probabilistic unary and binary predicates (e.g., \texttt{grass(x)}, \texttt{is\_flat\_surface(x)}, \texttt{near\_to(x,y)}), forming the exclusive interface to the symbolic reasoning engine. Because complexity scales with the number of regions and relations rather than image resolution, the abstraction enables efficient execution on embedded hardware.

\subsection{Rule-Based Reasoning Engine over the World Model}
\label{subsec:rules}

The decision layer is implemented in \textsc{Scallop}~\cite{huang2021scallop} as a probabilistic logic program over PSSG facts. Given unary and binary predicates grounded from perception and geometry, such as \texttt{grass(x)}, \texttt{is\_flat\_surface(x)}, and \texttt{near\_to(x,y)}, the rule program infers \texttt{hazard(x)} and \texttt{safe(x)} for candidate landing regions. The runtime program is fixed before evaluation. Thus, ``deterministic'' refers to deterministic execution of a fixed symbolic rule set over the current PSSG, although the input facts and proof weights may carry probabilities.

At runtime, no LLM is queried. The deployed system executes only the validated \textsc{Scallop} rule set over the current PSSG. During offline symbolic rule refinement, the LLM drafts candidate rules from expert-specified constraints and validation failures, and translates triggered rules into human-readable explanations for expert inspection. These draft programs are executed on validated PSSGs, iteratively refined through expert review, and included in the runtime rule set only after validation.


\subsubsection{Offline Human-in-the-Loop Rule Construction}
\label{subsec:hitl_offline}
Rule sets are synthesized offline using PSSGs generated from curated representative landing scenes~\cite{semanticDroneData}. Before rule synthesis, experts validate the PSSGs visually and programmatically by inspecting segmentation overlays, node attributes, and spatial relations. This separates perception and grounding errors from downstream reasoning errors.

In our implementation, OpenAI o3~\cite{openai2025chatgpt} drafts \textsc{Scallop} rules from the fixed PSSG schema, current rule library, and labeled landing examples. Candidate rules are executed over validated PSSGs in simulation, where the reasoning engine produces hazard and safety scores with provenance traces.

Experts then inspect predicted safe/unsafe regions, rule-trigger explanations, and mismatches between expected and inferred outcomes. When deficiencies are found, such as missing constraints, overly strict proximity thresholds, or unintended rule interactions, experts provide natural-language feedback. This feedback is converted into structured prompt refinements specifying missing predicates, threshold adjustments, or additional relational constraints. The process iterates until the generated rule set satisfies expert validation on the curated validation set.

This LLM-assisted process reduces the syntactic burden of writing \textsc{Scallop} programs while preserving expert control over the final decision logic. The accepted rule set remains explicit, editable, and inspectable. Only the symbolic logic is refined; the segmentation backbone and world-model construction remain fixed throughout.

\subsubsection{Rule notation}
We use Datalog-style notation for readability. 
In a rule \texttt{h(x) :- b1(x), b2(x,y)}, the symbol \texttt{:-} means that the head predicate \texttt{h(x)} is inferred when the body predicates hold. 
Commas denote conjunction, and \(\vee\) denotes disjunction; in the implemented \textsc{Scallop} program, disjunctive definitions are expanded into separate rules. 
Variables such as \(x\) and \(y\) range over PSSG nodes. 
Unary predicates such as \texttt{paved\_area(x)} are grounded from node semantics, while binary predicates such as \texttt{near\_to(x,y)} are grounded from deterministic spatial-relation construction. 
A candidate landing region satisfies \texttt{landable\_area(x)}; \texttt{hazard(x)} is then inferred to derive \texttt{safe(x)}.


\subsubsection{Runtime Execution over the World Model}

At runtime, the validated rule set executes directly over the online PSSG without any neural inference in the decision layer. \textsc{Scallop} evaluates the rule program as probabilistic first-order logic and answers queries such as $\mathrm{hazard}(r_i)$ and $\mathrm{safe}(r_i)$ for each candidate region $r_i$. The engine enumerates the top-$k$ weighted proofs and aggregates them into safety scores.

Because each proof is grounded in explicit predicates and rule clauses, provenance traces precisely record which world-model facts and logical rules contributed to each inference. This enables deterministic auditing, systematic debugging, and mission-level reconfiguration by directly modifying rule definitions rather than retraining neural components. A region is classified as safe if its safety score exceeds a mission-specific threshold.

\subsection{Multi-Frame Validation and Mission-Specific Ranking}
\label{subsec:mfv_ranking}

\paragraph{Multi-frame validation}
To mitigate transient segmentation noise, brief occlusions, and short-lived perception errors, candidate regions are validated over a short temporal window of $T$ consecutive frames. Let 
\[
W_T = \{t_0 - T + 1, \ldots, t_0\}
\]
denote the window ending at time $t_0$, and let $v^{(t)}$ denote the tracked instance of region $v$ at time $t$. A region passes multi-frame validation if:  
(i) it is never inferred as hazardous across the window, and  
(ii) its geometric properties remain stable. Formally,
\begin{equation}
\label{eq:mfv}
\mathrm{Pass}_T(v) \equiv 
\left( \bigwedge_{t \in W_T} \neg\,\mathrm{hazard}\!\left(v^{(t)}\right) \right)
\;\wedge\;
\neg\,\mathrm{Jitter}(v, W_T).
\end{equation}

Here, $\mathrm{hazard}(v^{(t)})$ is inferred by the symbolic reasoning engine at time $t$, and $\mathrm{Jitter}(\cdot)$ is a binary predicate that detects significant geometric instability (e.g., abrupt centroid or area variation). No additional threshold is introduced for jitter; a region either satisfies geometric consistency or it does not. We define the indicator
\[
I_T(v) = \mathbf{1}\{\mathrm{Pass}_T(v)\},
\]
which serves as a hard safety gate.

\paragraph{Mission-specific ranking}
Regions that satisfy $I_T(v)=1$ are admissible for landing. Final selection is performed via a mission-conditioned utility function. For each mission mode $m$, we compute normalized features $\tilde{b}_{m,k}(v) \in [0,1]$ (e.g., landing area, distance to target, estimated maneuver cost) together with nonnegative weights $\omega_{m,k}$ reflecting task priorities.

The mission score is defined as
\begin{equation}
\label{eq:ranking}
C_m(v) = I_T(v)\sum_k \omega_{m,k}\tilde{b}_{m,k}(v),
\qquad
v^\star = \arg\max_v C_m(v).
\end{equation}

Candidates failing multi-frame validation have \(I_T(v)=0\), with zero score, and are excluded from final selection. Importantly, \(C_m(v)\) is used solely as an inference-time ranking utility after symbolic safety filtering and is not used to train any model. The perception backbone is trained independently for semantic segmentation, while the \textsc{Scallop} rule set is constructed offline through rule synthesis and expert validation. For each mission mode, the weights \(\omega_{m,k}\) are predefined before evaluation and remain fixed across all test scenes.


\section{Experiments\label{sec:experiments}}
\subsection{Evaluation overview}
We evaluate \modelname{} along three axes: (i) effectiveness in software-in-the-loop (SIL) landing assessment, including an ablation that replaces the synthesized \textsc{Scallop} rules with deterministic FOL (\modelname-DetFOL); (ii) efficiency on edge hardware via HIL latency and resource profiling; and (iii) interpretability via qualitative case studies across mission modes (safe, emergency, rescue).

\subsection{Effectiveness in SIL}
We report \textbf{Succ} (successful assessments) and \textbf{MOD} (minimum obstacle distance) on successful trials.
Succ measures decision reliability under our protocol.
MOD is the minimum Euclidean distance from the predicted touchdown point to any annotated obstacle boundary; higher is safer:
$\mathrm{MOD}=\min_{\mathbf{o}\in\mathcal{O}}\|\hat{\mathbf{p}}-\mathbf{o}\|_2$.
We summarize MOD by median and P20 (20th percentile) over scenarios (5 runs each).



\noindent

\paragraph{Experiment Setup}
Experiments run in AirSim~\cite{shah2018airsim} on 72 scenarios under default conditions. Each baseline is executed 5 times per scenario.
We report Succ and MOD (median/P20).


\paragraph{Baselines}
All methods use monocular RGB and the same candidate representation where applicable.
We compare against SegOpticalFlow~\cite{Ho2018AdaptiveGainConstantFlowLanding,Scheper2020EvolutionOFLanding}, SafeUAV~\cite{marcu2018safeuav}, PEACE~\cite{Bong2025PEACE}, and a local VLM selector (LLMExplain, Phi-3.5-Vision)~\cite{abdin2024phi3,Phi35VisionInstruct2024}.
We also include \modelname-DetFOL, which replaces synthesized probabilistic rules with deterministic FOL over the same PSSG~\cite{Johnson2005HazardousTerrainLanding,Saripalli2003VisuallyGuidedLanding}.

\noindent
\paragraph{Mask-to-point conversion for evaluation}
For methods producing binary safe-region masks (SafeUAV and PEACE), the touchdown point is defined as the centroid of the largest predicted safe connected component. If no safe region is predicted (i.e., the output mask is empty), the prediction is treated as a failure.


\begin{table}[t]
\vspace{12pt}
\centering
\small
\renewcommand{\arraystretch}{1.25}
\setlength{\tabcolsep}{10pt}
\caption{Landing-site safety assessment on the AirSim test set (72 Scenarios).
\textbf{Succ.} is reported as \textbf{Safe/Land.}, where \textbf{Land.} denotes the number of scenarios in which the baseline produces at least one landing-site safety assessment decision (i.e., identifies one or more candidate sites), and \textbf{Safe} denotes the number of scenarios in which the predicted explicit touchdown point lies within the human-annotated safe landing zone.
\textbf{MOD} is reported in pixels as \textbf{median / 20th percentile (P20)}, where larger values indicate larger safety margins to surrounding obstacles.}
\label{tab:quality}
\resizebox{\columnwidth}{!}{%
\begin{tabular}{
l
@{\hspace{6pt}} c
@{\hspace{8pt}} c
}
\toprule
\textbf{Baseline} &
\makecell[c]{\textbf{Succ.$\uparrow$}\\\footnotesize Safe / Land.} &
\makecell[c]{\textbf{MOD (px)}$\uparrow$\\\footnotesize median / P20} \\
\midrule

\modelname              & \textbf{61/72} & 119.64 / 28.0 \\
\addlinespace[2pt]

SegOpticalFlow          & 53/63          & \textbf{159.40 / 51.19} \\
\addlinespace[2pt]

PEACE                   & 57/72          & 84.38 / 42.43 \\
\addlinespace[2pt]

SafeUAV                 & 47/72          & 74.48 / 18.36 \\
\addlinespace[2pt]

LLMExplain              & 37/72          & 52.00 / 13.0 \\
\addlinespace[4pt]

\modelname-DetFOL (Ablation) 
                        & 58/71          & 104.61 / 41.4 \\

\bottomrule
\end{tabular}

}
\end{table}

\begin{figure}[t]
    \centering
    \setlength{\tabcolsep}{2pt}
    
    \begin{subfigure}{0.48\columnwidth}
        \centering
        \includegraphics[width=\linewidth]{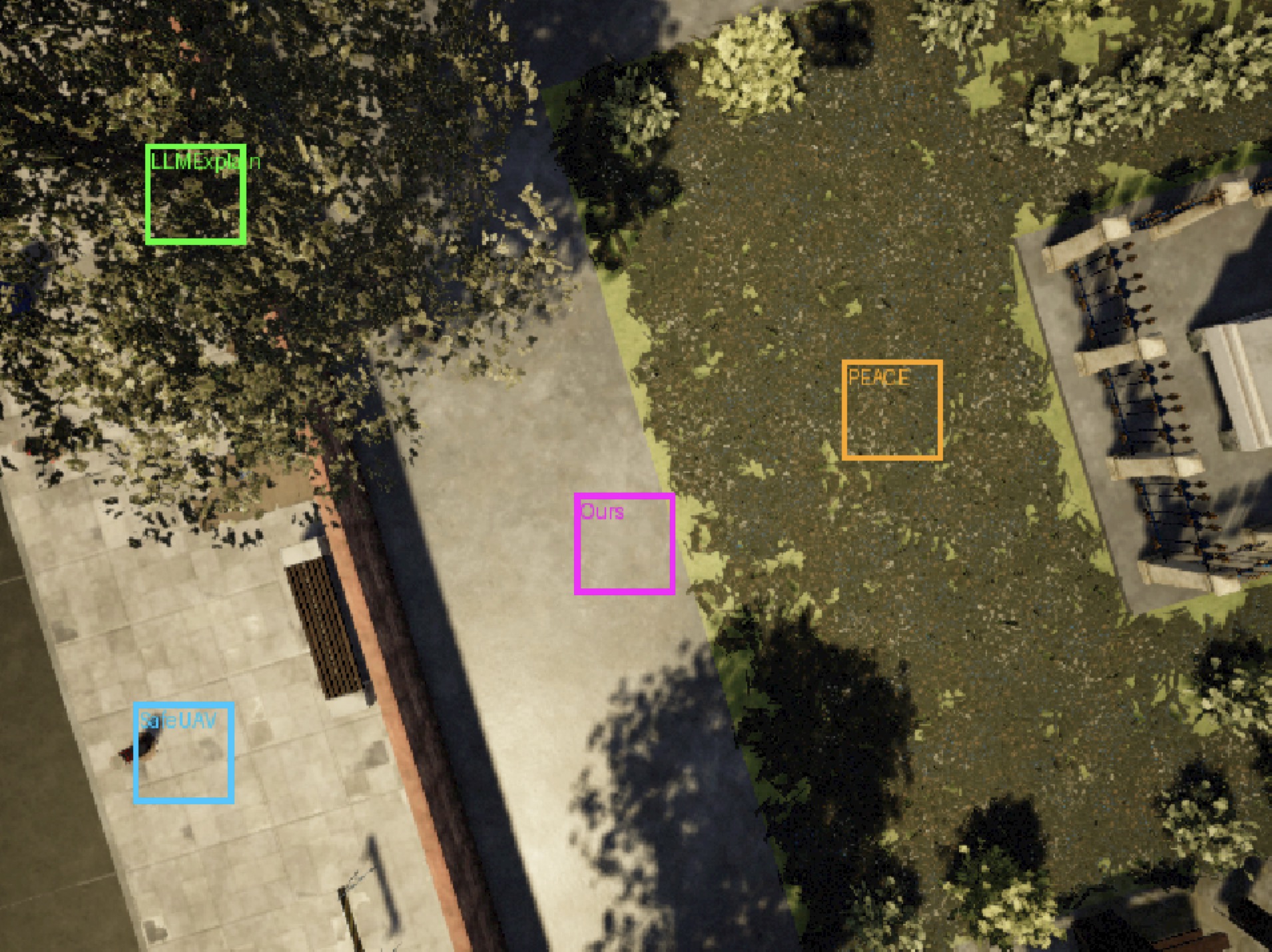}
        \caption{Success (RGB).}
        \label{fig:success_rgb}
    \end{subfigure}
    \hfill
    \begin{subfigure}{0.48\columnwidth}
        \centering
        \includegraphics[width=\linewidth]{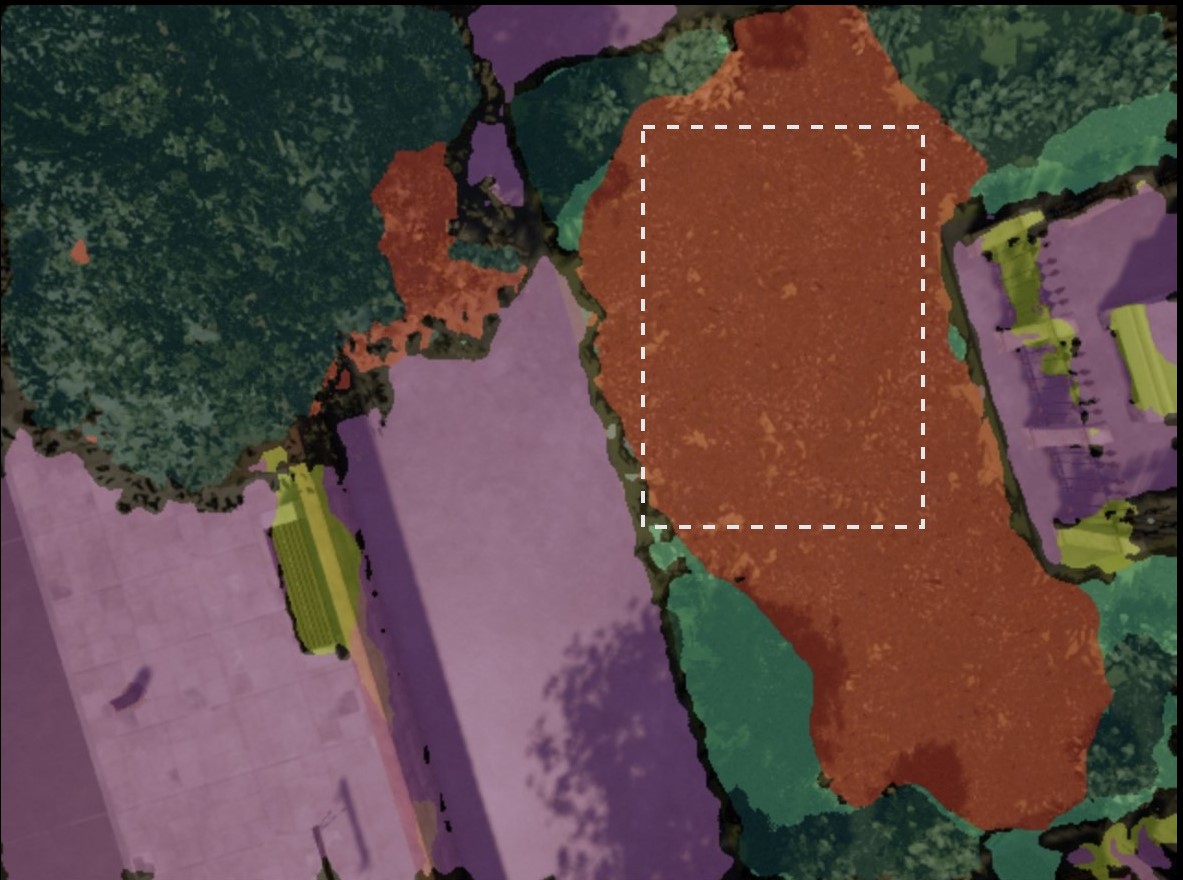}
        \caption{Segmentation.}
        \label{fig:success_seg}
    \end{subfigure}

    \vspace{5mm}

    \begin{subfigure}{0.48\columnwidth}
        \centering
        \includegraphics[width=\linewidth]{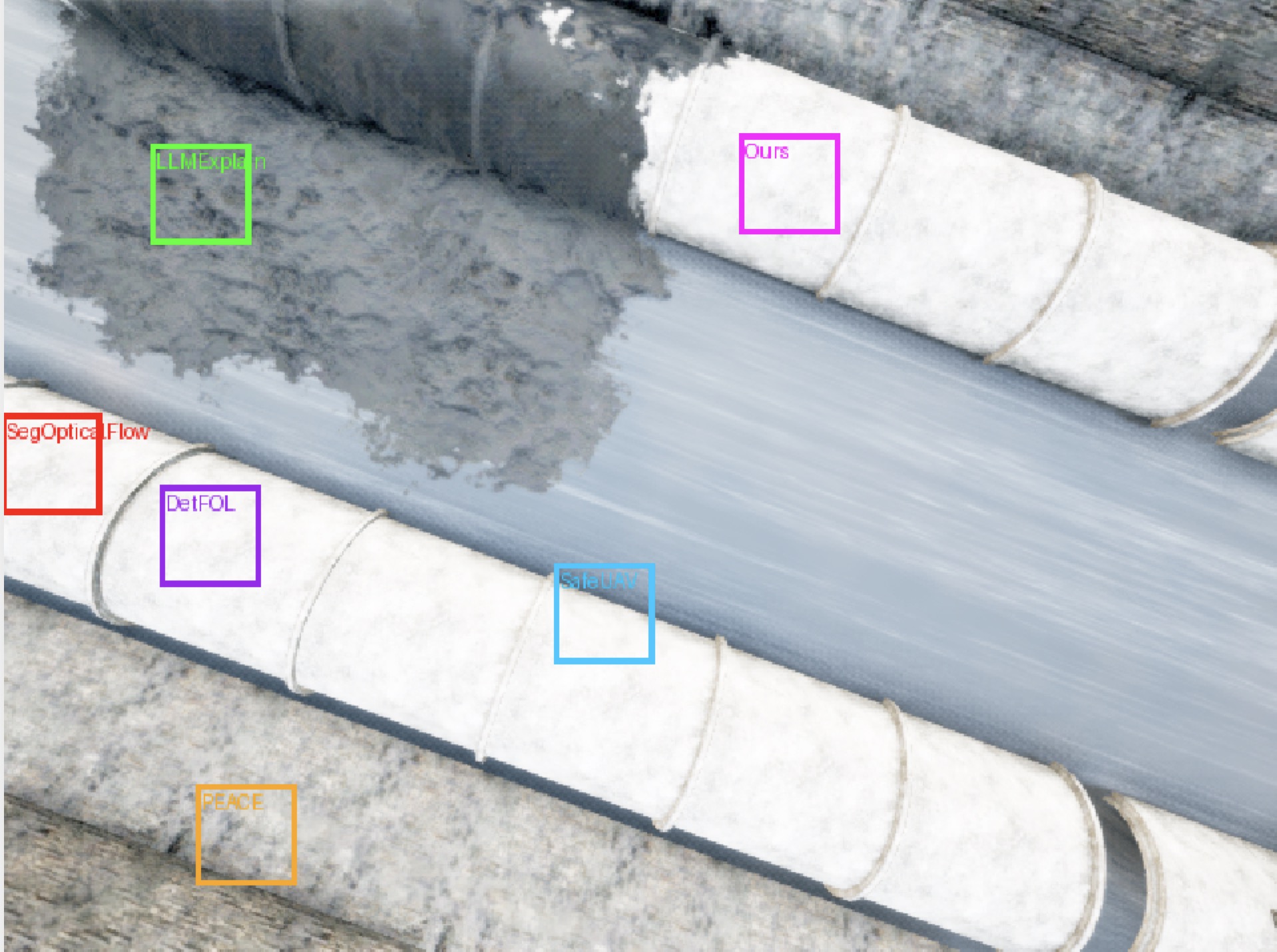}
        \caption{Perception-limited (RGB).}
        \label{fig:failure_rgb}
    \end{subfigure}
    \hfill
    \begin{subfigure}{0.48\columnwidth}
        \centering
        \includegraphics[width=\linewidth]{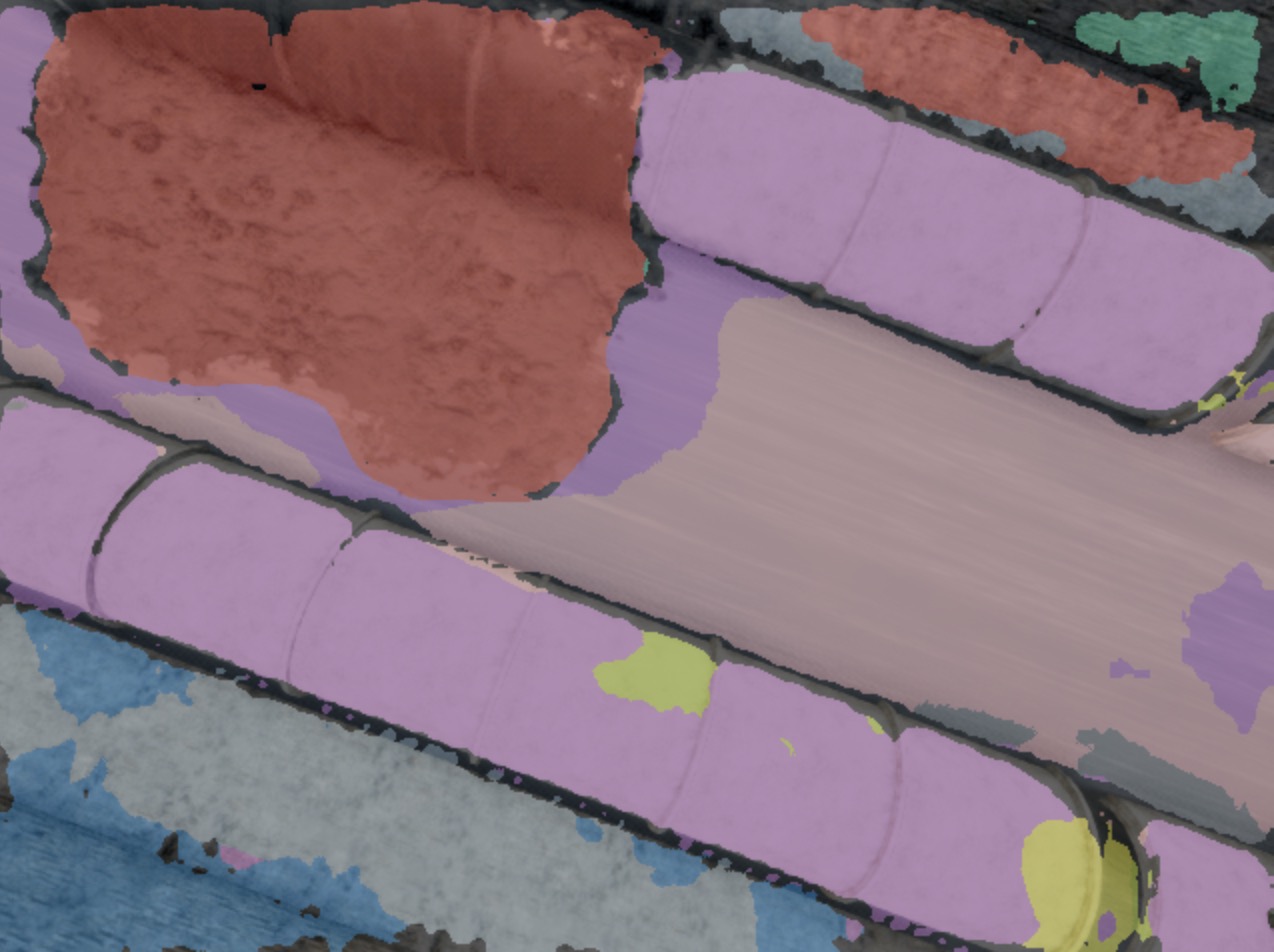}
        \caption{Segmentation.}
        \label{fig:failure_seg}
    \end{subfigure}
    \vspace{5mm}
\begin{subtable}[t]{0.45\linewidth}
  \centering
  \vspace{3mm}
  \begin{tabular}{@{} >{\centering\arraybackslash}p{1cm} l @{}}
    \toprule
    \textbf{Color} & \textbf{ Method} \\
    \midrule
    \cellcolor{green!80}  & LLMExplain \\
    \cellcolor{cyan!70} & SafeUAV \\
    \cellcolor{yellow!80}    & PEACE \\
    \cellcolor{red!80} & SegOpticalFlow \\
    \cellcolor{Fuchsia!70} & DetFOL \\
    \cellcolor{Rhodamine!70} & Ours (NeuroSymLand) \\
    \bottomrule
  \end{tabular}
  \caption{Color legend for landing-site assessment methods}
\end{subtable}%
\hfill
\begin{subtable}[t]{0.45\linewidth}
\vspace{3mm}
  \centering
  \begin{tabular}{@{} >{\centering\arraybackslash}p{1cm} l @{}}
    \toprule
    \textbf{Color} & \textbf{Semantic} \\
    \midrule
    \cellcolor{Mahogany!70}    & Water \\
    \cellcolor{yellow!80} & Obstacle \\
    \cellcolor{purple!50}   & Paved Area \\
    \cellcolor{ForestGreen!70} & Trees \\
    \cellcolor{pink!50} & Gravel \\
    \bottomrule
  \end{tabular}
  \caption{Color legend for semantic classes}
\end{subtable}


\caption{Qualitative landing-site assessment examples. \textbf{Top:} robust landing-site assessment despite fragmented semantic segmentation, where the same paved landing area is split into multiple disconnected regions and small isolated regions are incorrectly labeled (e.g., as water). \textbf{Bottom:} perception-limited failure caused by systematic semantic misclassification, where the paved landing area is  classified as gravel while the pipes are misclassified as paved area. 
}
    \label{fig:combined_cases}
\end{figure}

\paragraph{Result Analysis}
Table~\ref{tab:quality} shows that \modelname{} achieves the highest Succ score (61/72), outperforming PEACE (57/72) and SafeUAV (47/72). \modelname-DetFOL achieves 58/71, suggesting that explicit world modeling provides a strong foundation for landing-site assessment, while synthesized probabilistic rules further improve decision quality (+3 Succ). SegOpticalFlow reports the largest MOD, but it produces valid landing candidates in only 63 of 72 scenarios. Since MOD is undefined when no explicit touchdown point is produced, the reported clearance is computed only over the subset of scenarios with valid predictions. Therefore, its MOD should be interpreted together with the candidate-generation rate (63/72) and overall successful assessments (53/72). Inspection of the remaining failures suggests that most errors arise from perception failures that propagate incorrect predicates into the PSSG, consistent with the perception-limited example in Fig.~\ref{fig:combined_cases}.

\paragraph{Qualitative examples}
Fig.~\ref{fig:combined_cases} highlights how spurious semantic labels affect landing decisions and how explicit rules improve robustness.

\textbf{Case A (top): robustness under spurious labels and fragmented semantics.}
In this scene, the segmentation contains localized spurious regions (e.g., small patches marked by white dashed line predicted as \emph{water} near the boundary of otherwise landable surfaces), which fragments candidate masks and can perturb the touchdown point selected by purely mask- or score-driven pipelines.
Multiple methods still succeed here; however, \modelname{} remains stable because it reasons over region-level evidence in the PSSG and explicitly trades off geometric feasibility (flatness/area proxies), stand-off constraints (distance to obstacles/trees), and semantic vetoes.
In particular, the rule set penalizes candidates that are close to hazards even when a large flat surface exists, while allowing recovery from small isolated false positives via probabilistic grounding and relation-based reasoning (e.g., \texttt{near\_to}/\texttt{adjacent\_to}) rather than hard pixel exclusion.

\textbf{Case B (bottom): perception-limited failure due to systematic semantic swap.}
Here, a systematic misclassification swaps the semantics of the visually dominant surface and the pipes (e.g., the intended paved region becomes \emph{gravel} while the pipes are labeled as \emph{paved area}), corrupting predicate grounding in the world model.
Since all downstream logic operates on the explicit PSSG, the error propagates to candidate filtering and ranking, and no method can reliably recover without improved perception.
\subsection{Efficiency on edge hardware via HIL}\label{sec:hil}

\paragraph{HIL evaluation protocol and baselines}
For efficiency evaluation, we conduct 100 hardware-in-the-loop (HIL) trials on Jetson Orin Nano (8 GB) using recorded AirSim inputs. Initial poses and headings are randomized from a representative simulation map, capturing runtime variability while keeping environmental and task conditions controlled. For each trial, we measure end-to-end inference latency from image acquisition to landing-site safety output together with stage-wise execution time (semantic segmentation and post-processing, PSSG construction, symbolic reasoning, and mission-specific ranking). We additionally record system-level metrics including CPU/GPU utilization, memory footprint, power consumption, and operating temperature.

To contextualize latency and deployability, we compare \modelname{} against two lightweight baselines: SegOpticalFlow and SafeUAV, representing practical, industry-style markerless landing pipelines that can be deployed on the same embedded platform and execution stack. SegOpticalFlow reflects a perception-driven, motion-based embedded design, while SafeUAV represents a compact monocular learning-based approach. Heavier hybrid systems (e.g., PEACE, LLMExplain) are excluded from the HIL efficiency comparison because their foundation-model backbones exceed the memory and compute budget of the Jetson Orin Nano and consistently trigger out-of-memory (OOM) failures under our configuration. 

\paragraph{Results analysis}
All results are reported as \textbf{median $\pm$ standard deviation} over 100 HIL trials.

\begin{table}[t]
\vspace{14pt}
\centering
\caption{Stage-wise latency breakdown per frame.}
\label{tab:latency_breakdown}
\footnotesize
\begin{tabular}{lcc}
\toprule
Stage & Latency (ms) & Share \\
\midrule
Perception and post-processing & $424.4 \pm 30.7$ & $40.7\%$ \\
PSSG construction & $557.1 \pm 72.0$ & $53.4\%$ \\
Symbolic reasoning & $20.0 \pm 5.6$ & $1.9\%$ \\
Mission-specific ranking & $41.8 \pm 3.6$ & $4.0\%$ \\
\midrule
End-to-end & $1{,}043.3$ & $100\%$ \\
\bottomrule
\end{tabular}

\end{table}

\paragraph{Stage-wise latency breakdown}
As shown in Table~\ref{tab:latency_breakdown}, \modelname{} achieves an end-to-end latency of $1{,}043.3$~ms per frame ($\sim0.96$~FPS). The runtime is dominated by PSSG construction and perception/post-processing, which together account for over $94\%$ of the total latency. By contrast, symbolic reasoning takes only $20.0 \pm 5.6$~ms ($1.9\%$), showing that explicit rule-based inference introduces negligible overhead compared with perception and structured world-model construction.


\paragraph{System-level stability}
\modelname{} operates stably within embedded constraints, consuming $76.9\%$ CPU, $52\%$ GPU, $5.6$~GB memory, $12.3$~W power, and maintaining a temperature of $47.9^\circ$C. Compared with lightweight baselines, resource usage increases modestly due to explicit world-model construction but remains within the thermal and power envelope of the target platform.

\paragraph{Efficiency-robustness trade-off}
SafeUAV achieves $8.00$~FPS and SegOpticalFlow reaches $1.16$~FPS, whereas \modelname{} runs at $0.96$~FPS. While our approach incurs higher latency due to explicit graph construction and mission scoring, the refresh rate remains sufficient for the safety assessment, where update intervals on the order of one second are operationally acceptable. Importantly, Table~\ref{tab:quality} reveals that SafeUAV succeeds in only \textbf{47/72} scenarios and exhibits reduced MOD values, indicating smaller safety margins and weaker robustness in cluttered scenes. 

Taken together, the results indicate that \modelname{} achieves a favorable trade-off frontier: deployable on embedded hardware, negligible symbolic overhead, and substantially stronger robustness than perception-only baselines.

\begin{figure}[t]
\centering
\newcommand{\casewidth}{0.3\columnwidth}

\includegraphics[width=\casewidth]{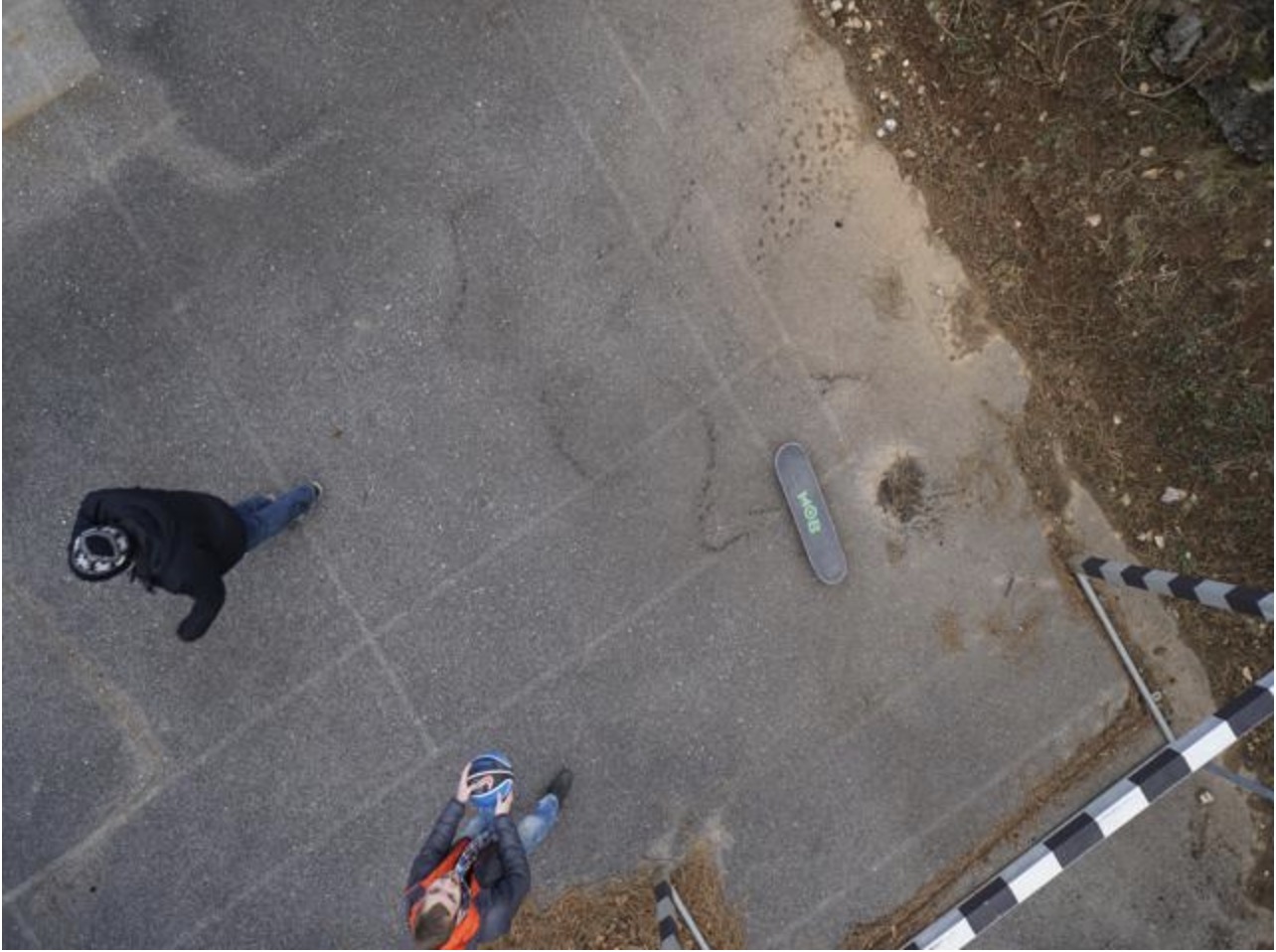}
\vspace{1pt}
\includegraphics[width=\casewidth]{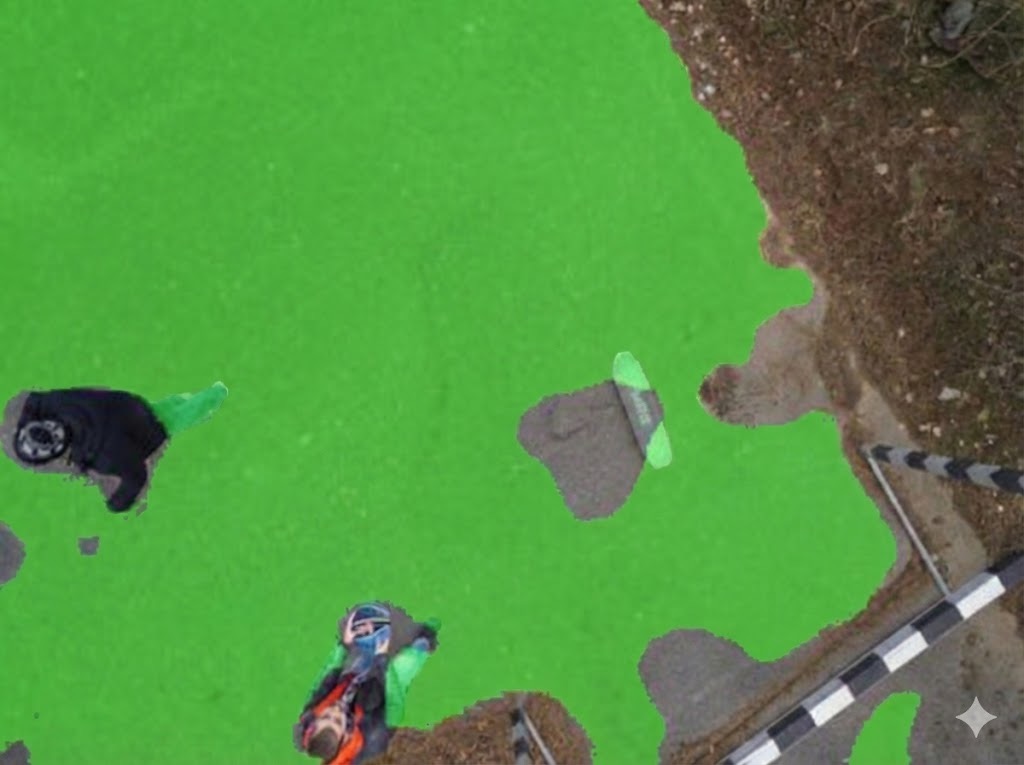}
\includegraphics[width=\casewidth]{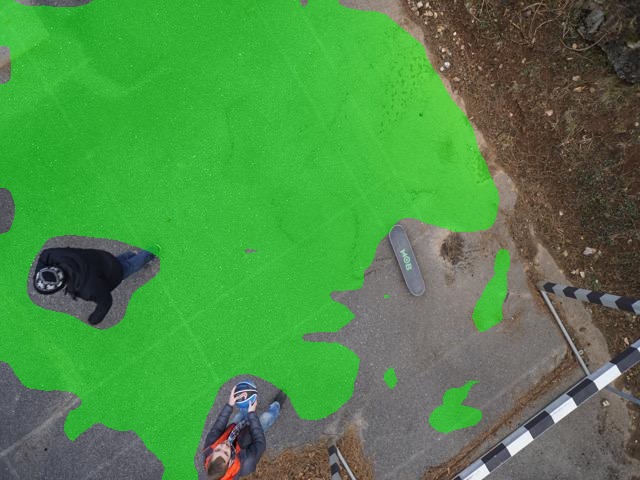}
\vspace{1pt}
\includegraphics[width=\casewidth]{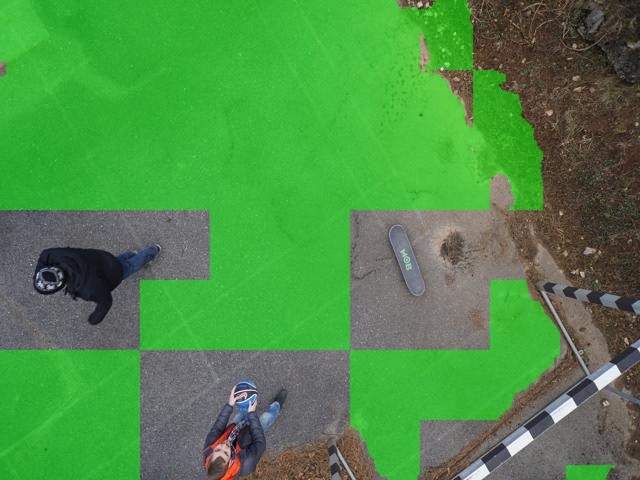}
\includegraphics[width=\casewidth]{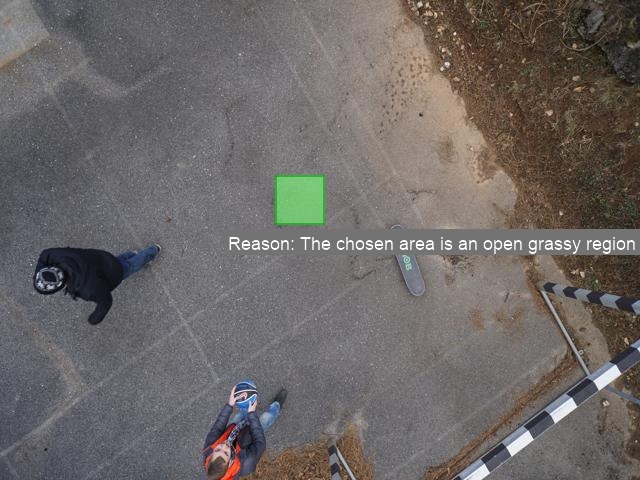}
\includegraphics[width=\casewidth]{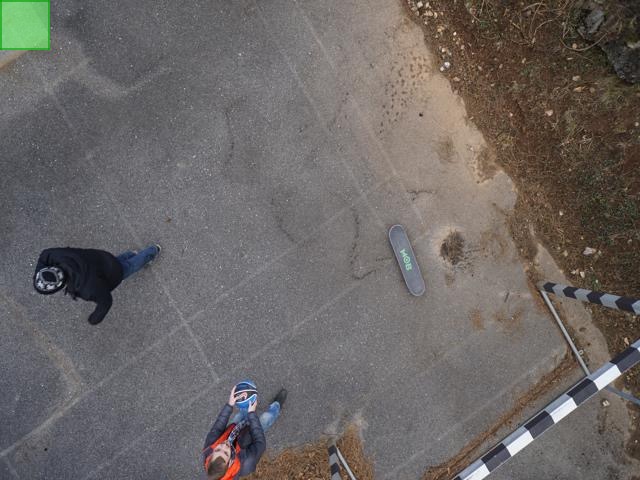}
\vspace{1pt}
\includegraphics[width=\casewidth]{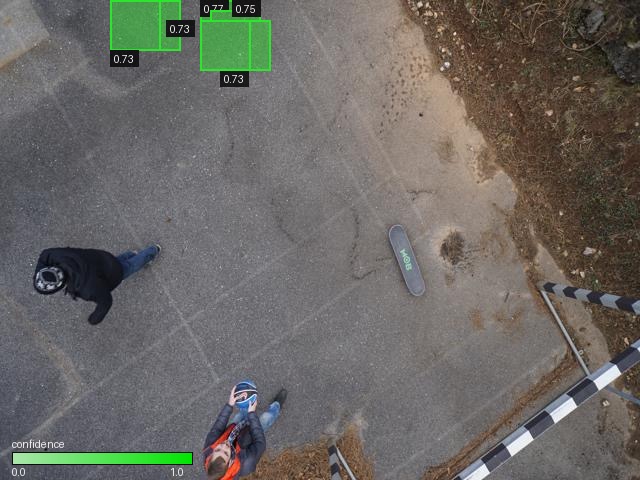}
\includegraphics[width=\casewidth]{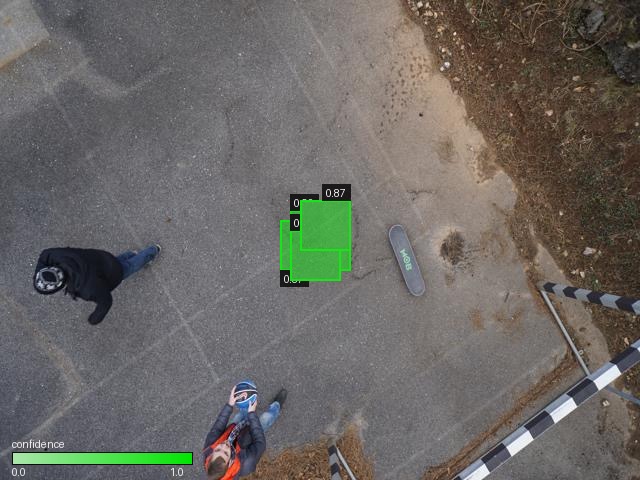}
\includegraphics[width=\casewidth]{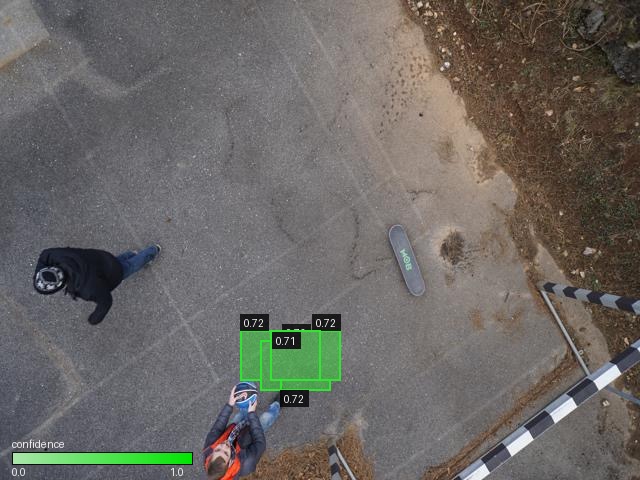}

\caption{Qualitative comparison on an AirSim scene.
Top row: RGB and learning-based baselines.
Middle: \modelname\ and alternative explainable baselines.
Bottom: mission-conditioned outputs (Safe-Landing/Emergency/Rescue).
\modelname\ integrates symbolic reasoning with mission context for task-adaptive decisions.}
\label{fig:rq3-case-study}
\end{figure}

\subsection{Interpretability and reasoning quality}
\label{sec:case-study}
We compare explanations and mission adaptation on a representative test scene (skateboard park).
Fig.~\ref{fig:rq3-case-study} visualizes baseline outputs and \modelname{} under three mission weightings (safe, emergency, rescue) with the same perception and PSSG held fixed.
We also report \textsc{Scallop} provenance statistics to attribute decisions to rule triggers grounded in scene-graph facts.

\paragraph{Results}
Fig.~\ref{fig:rq3-case-study} compares baseline outputs and \modelname{} on a representative scene under different mission contexts. SafeUAV and PEACE primarily produce dense safe-region masks, which identify large visually homogeneous paved areas but do not explicitly rank candidates by relational constraints such as stand-off distance to pedestrians, obstacles, or walls. LLMExplain outputs a single landing recommendation with a natural-language rationale, while SegOpticalFlow outputs a single landing point. Both provide limited access to intermediate safety evidence: LLMExplain is not grounded in executable spatial predicates, and SegOpticalFlow does not expose alternative candidates or mission-dependent trade-offs. As a result, these baselines are difficult to inspect or adapt when mission priorities change.

In contrast, \modelname{} evaluates candidate regions through explicit PSSG facts and \textsc{Scallop} rules. Under the default \textbf{Safe-Landing} context, candidates near people or obstacles are rejected through symbolic hazard rules such as \texttt{near\_person} and \texttt{near\_obstacle}. Under the \textbf{Emergency} context, ranking shifts toward proximity to the mission center while preserving the symbolic safety gate. Under the \textbf{Rescue} context, ranking prioritizes proximity to a designated target without allowing overlap with the target footprint. Across all three contexts, the perception pipeline, PSSG construction, and symbolic rule set remain unchanged; only the mission-conditioned ranking weights differ.

\paragraph{Rule-level analysis and interpretability}
\modelname{} provides provenance traces that attribute each decision to specific rules grounded in PSSG facts. In this case study, the hazard assessment is dominated by three proximity constraints: \texttt{r\_hazard\_near\_person} (mean proof weight $p{=}0.89$, triggered in 9 frames), \texttt{r\_hazard\_near\_obstacle} ($p{=}0.91$, 4 frames), and \texttt{r\_hazard\_near\_wall} ($p{=}0.97$, 4 frames). A vegetation containment rule, \texttt{r\_hazard\_cont\_veg}, triggers once ($p{=}0.99$), while other hazard rules remain inactive. These traces show that the selected landing decision is driven primarily by explicit stand-off constraints, making the rationale inspectable and adjustable by modifying symbolic predicates or relation thresholds without retraining perception.

\subsection{Discussion}
Our scope is \emph{landing-site assessment} rather than closed-loop landing execution. Accordingly, we evaluate the correctness of landing suitability assessment and the edge feasibility of the perception-to-reasoning pipeline through simulation and HIL experiments. The HIL results demonstrate that the assessment stack executes on the target embedded platform within bounded latency and resource constraints, rather than a complete onboard landing system. Integration with state estimation, trajectory planning, flight control, and real-world flight validation remains future work.

The curated rule-construction set, together with synthetic edge cases, is used solely for human-in-the-loop construction and refinement of the symbolic rule set, not for training the perception backbone. Experts review annotated landing outputs and concise rule-trigger summaries, providing high-level corrections when constraints are missing or overly restrictive through a small number of offline iterations. This keeps adaptation localized to the symbolic reasoning layer while preserving the separation between perception and decision logic. Evaluation is then performed on a disjoint test set with different scene compositions, ensuring the reasoning engine is assessed out-of-sample with respect to rule construction.



Current limitations mainly stem from perception quality and taxonomy coverage. Systematic segmentation errors can introduce incorrect facts into the PSSG, bounding the best achievable reasoning quality. In addition, our scenarios contain limited dynamics, such as moving obstacles, which restricts the benefit of richer temporal rules. Future work will extend evaluation to more dynamic scenes and broader taxonomies, incorporate complementary sensing modalities, and optimize PSSG construction through incremental graph updates for faster edge execution.

\section{Conclusion and Future Work}
\label{sec:conclusion}

We presented \modelname{}, a neuro-symbolic landing-site assessment framework that separates probabilistic world modeling from symbolic mission-level reasoning. An INT8-quantized segmentation backbone constructs an explicit PSSG, over which a \textsc{Scallop}-based rule engine performs interpretable safety reasoning with provenance traces. Mission-level safety behavior can be configured through human-in-the-loop construction and refinement of symbolic rules informed by representative scene images and expert constraints, without retraining the perception model. Across 72 simulated scenarios and 100 hardware-in-the-loop trials, \modelname{} improves landing-site assessment robustness over lightweight learning-based baselines while remaining deployable on embedded hardware.

The main limitation is perception-induced world-model errors: segmentation failures propagate incorrect predicates into symbolic reasoning, making reasoning fidelity fundamentally dependent on world-model quality. Future work will improve world-model reliability through uncertainty calibration, consistency checking, and region-level validation; extend reasoning to dynamic environments with temporal rules; incorporate complementary sensing modalities; optimize PSSG construction via incremental graph updates; and integrate the assessment module with onboard state estimation and flight control for closed-loop UAV landing.

\printbibliography
\end{document}